\documentclass[a4paper,conference]{IEEEtran}
\IEEEoverridecommandlockouts

\newcommand{\eg}{{\em e.g.}}

\newcommand{\Eq}[1]{Eq. (\ref{#1})}

\usepackage{color}
\definecolor{red}{rgb}{1.00,0.20,0.20}
\definecolor{blue}{rgb}{0.20,0.20,1.00}
\definecolor{green}{rgb}{0.00,1.00,0.00}

\usepackage{array}
\newcolumntype{L}[1]{>{\raggedright\let\newline\\\arraybackslash\hspace{0pt}}m{#1}}
\newcolumntype{C}[1]{>{\centering\let\newline\\\arraybackslash\hspace{0pt}}m{#1}}
\newcolumntype{R}[1]{>{\raggedleft\let\newline\\\arraybackslash\hspace{0pt}}m{#1}}

\newcommand{\UseShortAcronyms}{}

\ifdefined\UseShortAcronyms

\else

\fi

\usepackage[utf8]{inputenc}
\usepackage{comment}
\usepackage{amsfonts}
\usepackage{graphicx}
\usepackage{amsmath}
\usepackage{hyperref}

\ifCLASSINFOpdf
\else
\fi

\begin{document}
%
\title{Self-Supervised Domain Adaptation with Consistency Training}



%
\author{\IEEEauthorblockN{Liang Xiao\IEEEauthorrefmark{1} \thanks{This work is supported by National Natural Science Foundation of China under Grant No. 61803380 and 61790565.},
Jiaolong Xu\IEEEauthorrefmark{1},
Dawei Zhao\IEEEauthorrefmark{1},
Zhiyu Wang\IEEEauthorrefmark{2},
Li Wang\IEEEauthorrefmark{2},
Yiming Nie\IEEEauthorrefmark{1} and
Bin Dai\IEEEauthorrefmark{1}\IEEEauthorrefmark{2}}
}


\maketitle

\begin{abstract}
We consider the problem of unsupervised domain adaptation for image classification.
To learn target-domain-aware features from the unlabeled data, we create a self-supervised pretext task by augmenting the unlabeled data with a certain type of transformation (specifically, image rotation) and ask the learner to predict the properties of the transformation.
However, the obtained feature representation may contain a large amount of irrelevant information with respect to the main task. 
To provide further guidance, we force the feature representation of the augmented data to be consistent with that of the original data.
Intuitively, the consistency introduces additional constraints to representation learning, therefore, the learned representation is more likely to focus on the right information about the main task.
Our experimental results validate the proposed method and demonstrate state-of-the-art performance on classical domain adaptation benchmarks. Code is available at \url{https://github.com/Jiaolong/ss-da-consistency}.
\end{abstract}


%
\IEEEpeerreviewmaketitle

\section{Introduction}

Deep learning has shown remarkable performance in diverse machine learning problems and applications. However, the success of deep learning generally relies on a large amount of labeled data, which is not always available and often prohibitively expensive to acquire. To circumvent this issue, enormous efforts have been made, aiming at leveraging unlabeled data to improve the generalization performance. This line of research includes recent advances in transfer learning \cite{DAN:2015}, domain adaptation (DA) \cite{rot:2019}, semi-supervised \cite{uda:2019} and self-supervised learning \cite{revisiting:2019}.

In self-supervised learning, the model receives a supervision signal from an auxiliary task (also known as pretext task) without resorting to human annotations. The goal of self-supervised learning is to learn an useful feature representation for downstream tasks (such as image classification). A prominent subset of recent proposed auxiliary tasks have been evaluated in \cite{revisiting:2019}, such as predicting the context \cite{doersch:2015}, the relationship among different parts \cite{noroozi:2016}, and the transformation applied to the data \cite{gidaris:2018}. 
Among which image rotation prediction has been shown to yield good representation for image classification since
the orientation of an image is usually determined by the orientation of the object of interest.
On par with self-supervised learning, semi-supervised learning \cite{CSZ2006} is a classical paradigm to incorporate unlabeled data. 
A recent line of research \cite{uda:2019,hu2017learning} that exploits the consistency with respect to data augmentation has shown promising results on several benchmarks. 
The main idea is to ensure the predictions to be consistent before and after a perturbation/transformation of the input. 

\begin{figure}
	\centering
	\includegraphics[width=0.48\textwidth]{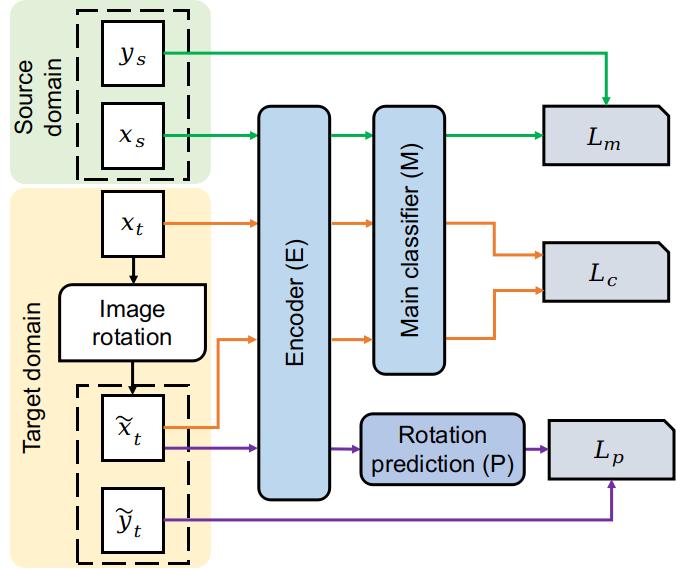}
	\caption{An overview of the proposed self-supervised domain adaptation with consistency training. 
		In the source domain, the labeled images are used to minimize the loss $\mathcal{L}_{m}$ of the main task.
		In the target domain, we have only unlabeled data, and thus augment the image $\textbf{x}_t$ to yield a rotated image $\tilde{\textbf{x}}_t$ and its corresponding rotation label $\tilde{y}_t$, which are then used to compute the consistency loss $\mathcal{L}_c$ and the pretext loss $\mathcal{L}_p$.
		All the losses will guide the update of the feature encoder to produce a domain invariant representation.
	}
	\label{fig:overview}
\end{figure}

Unsupervised domain adaptation can be viewed as a special case of semi-supervised learning, where the unlabeled data is drawn from a different data distribution due to the domain shift. 
In this case, a deep neural network learned from the source domain has no guarantee to generalize to the target domain \cite{CDAN:2018}.
A popular line of research for this problem builds on top of the generative adversarial network \cite{gan:2014}. 
Examples include domain adversarial neural network (DANN) \cite{Gani:2015} and its follow-ups \cite{DAN:2015,ADDA:2017,JAN:2017,CDAN:2018}.
However, due to the nature of the min-max optimization, adversarial DA methods are notoriously known to be unstable. Alternatively, the recent works by \cite{carlucci:2019} and \cite{rot:2019} introduce self-supervised learning 
to align the feature representations learned from different domains. 
Although self-supervised DA algorithms \cite{carlucci:2019,rot:2019} achieve the state-of-the-art results, 
the problem with self-supervised learning is that it may utilize undesired information to align different domains.
For example, \cite{Manasi06} shows that the image rotation can be predicted using only color moments computed from peripheral blocks in the image (i.e., mostly from the background), which is in general irrelevant to object recognition.

In this paper, 
we incorporate simultaneously the consistency loss to the self-supervised domain adaptation. 
Specifically, we overload the data augmentation operation, such as image rotation, to bridge self-supervised learning and consistency learning.
An overview of our method is illustrated in Figure \ref{fig:overview}. 
Our intuition is that the consistency loss potentially reduces the chance of relying on non-robust features which are discriminative for predicting rotations but are not important for the main task. As such, the learned feature representation will more likely to serve the main task. 
From an information theoretical point of view, 
incorporating consistency learning amounts to maximizing the mutual information between the transformed data and the label of the main task.
This plays a critical regularization on how we extract information from the transformed/augmented data.
We validate the proposed method on standard DA benchmarks, such as the Office-31, PACS, ImageCLEF and Office-Home datasets,
and show that it constantly outperforms self-supervised DA methods and other state-of-the-art DA methods.


Our main contributions are summarized as follows:

\begin{itemize}

\item We propose a new domain adaptation method by leveraging the fact that the data augmentation operation can be shared simultaneously by self-supervised learning and consistency learning. 

\item We derive an information theoretical understanding of the consistency loss and justify it is complementary to self-supervised learning. 

\item We show that the proposed method achieves state-of-the-art performance on several domain adaptation benchmarks.
\end{itemize}

\vspace{-1mm}
\section{Related work}

Our work is related to self-supervised representation learning, consistency training for semi-supervised learning and domain adaptation. For self-supervised representation learning, the design of the self-supervised learning task has been seen crucial to the accuracy of the downstream task \cite{revisiting:2019}. Various pretext tasks have been proposed, including image colorization \cite{colorful:2016}, image in-painting \cite{pathak:2016}, patch-level relative position prediction \cite{doersch:2015} \cite{jigsaw:2018}  and image rotation prediction \cite{gidaris:2018}. According to a recent review \cite{revisiting:2019}, the simple image rotation prediction pretext task has shown promising results. The image rotation prediction has also been successfully applied to address domain adaptation problem in \cite{rot:2019}, where it also outperforms other counterparts, {\eg}, jigsaw puzzle based methods \cite{jigsaw:2018} \cite{carlucci:2019}. 

The consistency training has been seen popular in recent semi-supervised learning literature. Different consistency training methods vary in how to generate data perturbation and how the consistency loss is composed. For visual learning, composition of translation, flipping, rotation, stretching, shearing and adding noise is often adopted \cite{Dosovitskiy:2014}. In \cite{mixup:2018}, a method called MixUp is proposed, which performs linear interpolation between the samples to generate
virtual samples. In \cite{uda:2019}, the influence of the form of the noise to consistency training is investigated. Apart from these manually designed data augmentation,  virtual adversarial training (VAT) \cite{vat:2018} searches for perturbation that maximizes the change in the prediction of the model. Policy search based augmentation approaches have also been proposed, such as AutoAugment \cite{AutoAugment:2019}, Population based augmentation (PBA) \cite{pba:2019} and RandAugment \cite{randaugment:2019}. The work of \cite{French:2018} uses consistent prediction as a constraint and adapts a mean teacher model for domain adaptation. As far as we know, this is the most relevant work of applying consistency training with domain adaptation.

For domain adaptation, the most related work to ours is the self-supervised DA of \cite{rot:2019}, where both image rotation prediction and jigsaw puzzle based pretext tasks are investigated. Base on the conclusion of \cite{rot:2019}, we focus on image rotation prediction, and propose a consistency training method which constantly improves the accuracy of self-supervised DA.

\vspace{-1mm}
\section{Method}

In this section, we first revisit the self-supervised domain adaptation. On top of that, the consistency training is proposed to improve the representation learning with domain invariance properties. For the sake of clarity, through out this paper, we focus on image classification problems even though our method is generic and can be applied for other tasks. 
\subsection{Self-supervised domain adaptation}
\label{sec:ssl_da}
We denote the source domain by $\mathit{D}_s = \{\textbf{x}^s_i, y^s_i\}_{i=0}^{N_s}$, where $\textbf{x}^s_i$ and $\textbf{y}^s_i$ are the image and corresponding class label respectively. We denote the unlabeled target domain by $\mathit{D}_t = \{\textbf{x}^t_i\}_{i=0}^{N_t}$. The convolutional neural network (CNN) is used for classification, which can be decomposed as a feature encoder and a linear classifier. 
We use $E$ to denote the encoder network and $M$ to denote the classification network. 
The parameters of $E$ and $M$ are denoted by $\boldsymbol{\theta_{e}}$ and $\boldsymbol{\theta_{m}}$ respectively. For self-supervised DA, a new branch $P$ with parameter $\boldsymbol{\theta_{p}}$ is added to the output layer of $E$ for self-supervised learning. The combination of $E$, $M$ and $P$ is essentially a multi-head CNN. For the choice of the pretext task of branch $P$, we use image rotation prediction as in \cite{rot:2019}. The rotation operation is defined by 
$\tilde{\textbf{x}} = \mathrm{Rot}(\textbf{x},\tilde{y}), \tilde{y} \in [0, 3]$, 
which rotates image $\textbf{x}$ by $\tilde{y} \cdot 90$ degrees. We denote by $\tilde{\mathit{D}}_t = \{\tilde{\textbf{x}}^t_i, \tilde{y}^t_i\}_{i=0}^{N_t} $ the set of rotated target domain examples. The image rotation prediction model $\mathit{P}$ takes as input the feature representation from $\mathit{E}$ and outputs the probability of each rotation angle. The self-supervised DA solves the following multi-task learning objective:
\begin{align}
	\label{eq:obj_rot}
	\min_{\boldsymbol{\theta_{e}}, \boldsymbol{\theta_{m}}, \boldsymbol{\theta_{p}}} &\mathcal{L}_{m}(\boldsymbol{\theta_{e}}, \boldsymbol{\theta_{m}}) + \lambda_p \mathcal{L}_{p}(\boldsymbol{\theta_{e}}, \boldsymbol{\theta_{p}}), \\
	&\mathcal{L}_{m}(\boldsymbol{\theta_{e}}, \boldsymbol{\theta_{m}}) = \mathbb{E}_{\textbf{x}^s, y^s \in \mathit{D}_s}\big[-\log{p(y^s \mid \textbf{x}^s)}\big], \\
	&\mathcal{L}_{p}(\boldsymbol{\theta_{e}}, \boldsymbol{\theta_{p}}) = 
	\mathbb{E}_{\textbf{x}^t \in \mathit{D}_t}\big[-\log{p(\tilde{y}^t \mid \textbf{x}^t)}\big],
\end{align}
where $\mathcal{L}_{m}$ and $\mathcal{L}_{p}$ are the cross-entropy loss functions for the main classification task and the pretext task respectively; $\lambda_p$ is the coefficient controlling the influence of the pretext task.
Note that both $p(y^t \mid \textbf{x}^t)$ and $p(\tilde{y}^t \mid \textbf{x}^t)$ depend on the feature encoder $E$.

\subsection{Self-supervised DA with consistency training}


The pretext task in self-supervised learning is designed to be compatible with the main task of interest.
However, most of these designs are based on intuition or heuristics. 
There is no guarantee on the compatibility between the pretext task and the main task.
For example, the prediction of image rotation can perform well with only the low-level features extracted from the peripheral area in the image \cite{Manasi06}.
If the feature encoder cheats by reserving a part of the representation to capture these low-level features,
the self-supervision loss can still be minimized, but it is not necessarily a good representation for the main task.
We thus would like to explicitly relate the representation of the rotated image $\tilde{\textbf{x}}$ to the label $y$ of the main task.
To this end, we propose to explicitly maximize the mutual information between
the rotated image and the label. Formally, the negative mutual information is given by

\newcommand{\thinmid}{\,|\,} 
\begin{align}
	&-I(\tilde{\textbf{x}}; y) = -\mathbb{E}_{p(\textbf{x},\tilde{\textbf{x}},y)} \Big[ \log \frac{p(y \mid \tilde{\textbf{x}})}{p(y)} \Big] \\
	&= \mathbb{E}_{p(\textbf{x},\tilde{\textbf{x}})}\Big[\! - \!\sum_y p(y \thinmid \textbf{x}) \log \frac{p(y \thinmid \tilde{\textbf{x}})}{p(y)}
	+ \sum_y p(y \thinmid \textbf{x}) \log \frac{p(y \thinmid \textbf{x})}{p(y \thinmid \textbf{x})} \Big] \\
	&= \mathbb{E}_{p(\textbf{x},\tilde{\textbf{x}})}\Big[ \sum_y p(y \thinmid \textbf{x}) \log \frac{p(y \thinmid \textbf{x})}{p(y \thinmid \tilde{\textbf{x}})}
	- \sum_y p(y \thinmid \textbf{x}) \log \frac{p(y \thinmid \textbf{x})}{p(y)} \Big] \\
	&= \mathbb{E}_{p(\textbf{x},\tilde{\textbf{x}})}\Big[ \mathcal{D}_\text{KL}\big(p(y \thinmid \textbf{x}) \| p(y \mid \tilde{\textbf{x}})\big)
	- \mathcal{D}_\text{KL}\big(p(y \mid \textbf{x}) \| p(y)\big) \Big],
 \label{eq:mi_final}
\end{align}
where the joint distribution $p(y, \textbf{x},\tilde{\textbf{x}}) = p(\textbf{x},\tilde{\textbf{x}}) p(y \mid \textbf{x})$ with an assumption that $y$ is conditionally independent of $\tilde{\textbf{x}}$ given $\textbf{x}$.
If we assume that $p(y)$ is uniform, then the second term in eq.\eqref{eq:mi_final} simplifies as $\mathbb{E}_{\textbf{x}}[-\sum_y p(y \mid \textbf{x}) \log p(y \mid \textbf{x})]$. 

The decomposition of $-I(\tilde{\textbf{x}}; y)$ suggests adding two terms to the objective function of self-supervised DA, which turns out to be well-known losses in the literature:
\begin{align}
	\label{eq:loss_rot_c}
	&\mathcal{L}_{c}(\boldsymbol{\theta_{e}}, \boldsymbol{\theta_{m}}) = \mathbb{E}_{\textbf{x}^t \in \mathit{D}_t} \mathbb{E}_{\tilde{\textbf{x}}^t \in \tilde{\mathit{D}}_t} \Big[\mathcal{D}_\text{KL}(\hat{p}(y^t \mid \textbf{x}^t) \| p(y^t \mid \tilde{\textbf{x}^t}))\Big], \\
	\label{eq:loss_ent_min}
	&\mathcal{L}_{e}(\boldsymbol{\theta_{e}}, \boldsymbol{\theta_{m}}) = \mathbb{E}_{\textbf{x}^t \in \mathit{D}_t }\Big[-\sum_{y^t} p(y^t \mid \textbf{x}^t)\log{p(y^t \mid \textbf{x}^t)}\Big],
\end{align}
where $\mathcal{L}_{c}(\boldsymbol{\theta_{e}}, \boldsymbol{\theta_{m}})$ is known as the Kullback-Leibler consistency loss as considered by VAT \cite{vat:2018} and UDA \cite{uda:2019} for semi-supervised learning;
$\mathcal{L}_{e}(\boldsymbol{\theta_{e}}, \boldsymbol{\theta_{m}})$ is known as the entropy minimization loss (EntMin) \cite{EntMin:2005}.
Note that $\hat{p}(y^t \mid \textbf{x}^t)$ is a copy of $p(y^t \mid \textbf{x}^t)$ with fixed $\boldsymbol{\theta_{e}}$ and $\boldsymbol{\theta_{m}}$ such that no gradients back-propagate to them.

Intuitively, the consistency loss forces the representation to be insensitive to data augmentation. 
However, this does not imply that the consistency loss is in conflict with the self-supervision loss.
We would rather view them as two different constraints to representation learning.
Recall that the ideal representation of the objects of interest satisfies both constraints.
Similarly, the EntMin loss imposes the third constraint that penalizes vague predictions,
leading to more discriminative representations.
In addition, $\hat{p}(y^t \mid \textbf{x}^t)$ can be viewed as the pseudo label, 
which gradually propagates the supervision signal from the labeled source domain to the unlabeled target domain.

The final objective function is thus
\begin{align}
	\label{eq:obj_rot_c2}
	\min_{\boldsymbol{\theta_{e}}, \boldsymbol{\theta_{m}}, \boldsymbol{\theta_{p}}} &\mathcal{L}_{m}(\boldsymbol{\theta_{e}}, \boldsymbol{\theta_{m}}) + \lambda_p \mathcal{L}_{p}(\boldsymbol{\theta_{e}}, \boldsymbol{\theta_{p}})  \notag \\
	&\lambda_c \mathcal{L}_{c}(\boldsymbol{\theta_{e}}, \boldsymbol{\theta_{m}}) + \lambda_e \mathcal{L}_{e}(\boldsymbol{\theta_{e}}, \boldsymbol{\theta_{m}}), 
\end{align}
where $\lambda_c, \lambda_e$ are respectively coefficients for the consistency loss and the EntMin loss.

\section{Experiments and results}

\label{sec:experiment}

In this section, we conduct experiments to evaluate the proposed domain adaptation method and compare the results with state-of-the-art methods. 

\subsection{Datasets and setup}

We implement the proposed method using the PyTorch framework on a single GTX 1080 Ti GPU with $11$ GB memory. ResNet-18 and ResNet-50 architectures are used as base networks and initialized with ImageNet \cite{Deng:2009} pretrained weights. We evaluate on the following popular domain adaptation datasets:

\textbf{PACS} \cite{pacs:2017} has $7$ object categories and $4$ domains (Photo, Art Paintings, Cartoon and Sketches). We follow the multi-source domain adaptation settings as in \cite{carlucci:2019} and \cite{rot:2019} and trained our model considering three domains as source datasets and the remaining one as target.

\textbf{Office-31} \cite{Saenko:2010} is the most widely used dataset for visual domain adaptation, which has $4652$ images and $31$ categories collected from three domains: Amazon (\textbf{A}), Webcam (\textbf{W}) and DSLR (\textbf{D}). We evaluate on six domain adaptation tasks, $A \rightarrow W$, $D \rightarrow W$, $A \rightarrow D$, $D \rightarrow A$ and $W \rightarrow A$.

\textbf{ImageCLEF} \cite{Saenko:2010} consists of three domains, including Caltech-256 (C), ImageNet 2012 (I), and Pascal VOC 2012 (P). The three domains share 12 common classes. Six domain adaptation tasks are evaluated on ImageCLEF: $I \rightarrow P$, $P \rightarrow I$, $I \rightarrow C$, $C \rightarrow I$, $C \rightarrow P$ and $P \rightarrow C$.

\textbf{Office-Home} \cite{deephash:2017} contains 4 domains, each domain consists of images from 65 categories of everyday objects. The total number of the images is about 15,500. The 4 domains are: Art (\textbf{ar}), Clipart (\textbf{cl}), Product (\textbf{pr}) and Real-World (\textbf{rw}). We evaluate on the domain adaptation tasks on each two domains of Office-Home dataset.

Following the standard protocols of unsupervised domain adaptation \cite{Gani:2015} and \cite{JAN:2017}, we use all labeled source domain examples and all unlabeled target domain examples. We set three different random seeds and run each experiment three times. The final result is the average over the three repetitions. We compare our proposed method with state-of-the-art DA methods: \textbf{Dial} \cite{carlucci:2017}, \textbf{DDiscovery} \cite{mancini:2018}, 
\textbf{DAN} \cite{DAN:2015}, \textbf{DANN} \cite{Gani:2015}, \textbf{RTN} \cite{RTN:2016}, \textbf{ADDA} \cite{ADDA:2017}, \textbf{JAN} \cite{JAN:2017}, \textbf{CDAN} \cite{CDAN:2018}, \textbf{JiGen} \cite{carlucci:2019}, \textbf{Jigsaw} \cite{rot:2019} and \textbf{Rot} \cite{rot:2019}.

\subsection{Results}

\begin{table}
	\center
	\caption{Multi-source Domain Adaptation results on PACS (ResNet-18). Three domains are used as source datasets and the remaining one as target.}
	\label{tab:pacs}
	\begin{tabular}[htb]{cccccc}
\hline
Method & Art. & Cartoon & Sketch & Photo & Avg. \\ \hline
SRC & 74.7 & 72.4 & 60.1 & 92.9 & 75.0 \\
Dial & 87.3 & 85.5 & 66.8 & 97.0 & 84.2 \\
DDiscovery & 87.7 & 86.9 & 69.6 & 97.0 & 85.3 \\ \hline
CDAN & 85.7 & 88.1 & 73.1 & 97.2 & 86.0 \\ 
CDAN+E & 87.4 & \textbf{89.4} & \textbf{75.3} & 97.8 & 87.5 \\ \hline
JiGen & 84.9 & 81.1 & 79.1 & 97.9 & 85.7 \\
Jigsaw & 84.9 & 83.9 & 69.0 & 93.9 & 82.9 \\
Rot & 88.7 & 86.4 & 74.9 & \textbf{98.0} & 87.0 \\ \hline
Ours & \textbf{90.3} & 87.4 & 75.1 & 97.9 & \textbf{87.7} \\
\hline
\end{tabular}
\end{table}

\begin{table*}
	\center
	\caption{Accuracy (\%) on Office-31 dataset (ResNet-50).}
	\label{tab:office}
	\resizebox{\textwidth}{!}{
\begin{tabular}[htb]{cccccccc}
\hline
Method & ${A}\rightarrow{W}$ & ${D}\rightarrow{W}$ & ${W}\rightarrow{D}$ & ${A}\rightarrow{D}$ &  ${D}\rightarrow{A}$ & ${W}\rightarrow{A}$ & Avg. \\ \hline
ResNet-50  &68.4$\pm$0.2 &96.7$\pm$0.1 &99.3$\pm$0.1 &68.9$\pm$0.2 &62.5$\pm$0.3 &60.7$\pm$0.3 &76.1 \\
DAN  &80.5$\pm$0.4 &97.1$\pm$0.2 &99.6$\pm$0.1 &78.6$\pm$0.2 &63.6$\pm$0.3 &62.8$\pm$0.2 &80.4 \\
RTN  &84.5$\pm$0.2 &96.8$\pm$0.1 &99.4$\pm$0.1 &77.5$\pm$0.3 &66.2$\pm$0.2 &64.8$\pm$0.3 &81.6 \\
DANN  &82.0$\pm$0.4 &96.9$\pm$0.2 &99.1$\pm$0.1 &79.7$\pm$0.4 &68.2$\pm$0.4 &67.4$\pm$0.5 &82.2 \\
ADDA  &86.2$\pm$0.5 &96.2$\pm$0.3 &98.4$\pm$0.3 &77.8$\pm$0.3 &69.5$\pm$0.4 &68.9$\pm$0.5 &82.9 \\
JAN  &85.4$\pm$0.3 &97.4$\pm$0.2 &99.8$\pm$0.2 &84.7$\pm$0.3 &68.6$\pm$0.3 &70.0$\pm$0.4 &84.3 \\
CDAN  &93.1$\pm$0.2 &98.2$\pm$0.2 &\textbf{100.0}$\pm$0.0 &89.8$\pm$0.3 &70.1$\pm$0.4 &68.0$\pm$0.4 &86.6 \\
CDAN+E  &\textbf{94.1}$\pm$0.1 &\textbf{98.6}$\pm$0.1 &\textbf{100.0}$\pm$0.0 &\textbf{92.9}$\pm$0.2 &71.0$\pm$0.3 &69.3$\pm$0.3 &\textbf{87.7} \\ \hline
Jigsaw &86.9$\pm$0.8 &\textbf{98.6}$\pm$0.5 &\textbf{100.0}$\pm$0.0 &82.9$\pm$1.0 &62.9$\pm$1.2 &61.2$\pm$0.7 &82.1
\\
Rot &90.1$\pm$0.8 &98.1$\pm$0.3 &\textbf{100.0}$\pm$0.0 &88.6$\pm$0.7 &65.1$\pm$0.8 &65.0$\pm$0.6 &84.5 \\ \hline
Ours &92.5$\pm$0.2 &\textbf{98.7}$\pm$0.3 &\textbf{100.0}$\pm$0.0 &88.6$\pm$0.2 &69.4$\pm$0.4 &67.2$\pm$0.3 &86.1 \\
\hline
\end{tabular}
}
\end{table*}

\begin{table*}
	\center
	\caption{Accuracy (\%) on Image-CLEF (ResNet-50).}
	\label{tab:image_clef}
	\resizebox{\textwidth}{!}{
\begin{tabular}[htb]{cccccccc}
\hline
Method & ${I}\rightarrow{P}$ & ${P}\rightarrow{I}$ & ${I}\rightarrow{C}$ & ${C}\rightarrow{I}$ &  ${C}\rightarrow{P}$ & ${P}\rightarrow{C}$ & Avg. \\ \hline
ResNet-50 &74.8$\pm$0.3 &83.9$\pm$0.1 &91.5$\pm$0.3 &78.0$\pm$0.2 &65.5$\pm$0.3 &91.2$\pm$0.3 &80.7 \\
DAN & 74.5$\pm$0.4 &82.2$\pm$0.2 &92.8$\pm$0.2 &86.3$\pm$0.4 &69.2$\pm$0.4 &89.8$\pm$0.4 &82.5 \\
DANN & 75.0$\pm$0.6 &86.0$\pm$0.3 &96.2$\pm$0.4 &87.0$\pm$0.5 &74.3$\pm$0.5 &91.5$\pm$0.6 &85.0 \\
JAN & 76.8$\pm$0.4 &88.0$\pm$0.2 &94.7$\pm$0.2 &89.5$\pm$0.3 &74.2$\pm$0.3 &91.7$\pm$0.3 &85.8 \\
CDAN & 76.7$\pm$0.3 & 90.6$\pm$0.3 & 97.0$\pm$0.4 & 90.5$\pm$0.4 & \textbf{74.5}$\pm$0.3 & 93.5$\pm$0.4 & 87.1 \\
CDAN+E & 77.7$\pm$0.3 & 90.7$\pm$0.2 & \textbf{97.7}$\pm$0.3 & \textbf{91.3}$\pm$0.3 & 74.2$\pm$0.2 & 94.3$\pm$0.3 & \textbf{87.7} \\
Rot &77.9$\pm$0.8 &91.6$\pm$0.3 &95.6$\pm$0.2 &86.9$\pm$0.6 &70.5$\pm$0.7 &94.8$\pm$0.3 &84.2 \\ \hline
Ours &\textbf{78.6}$\pm$0.4 &\textbf{92.5}$\pm$0.1 &96.1$\pm$0.3 &88.9$\pm$0.2 &73.9$\pm$0.7 &\textbf{95.9}$\pm$0.6 &\textbf{87.7} \\
\hline
\end{tabular}
}
\end{table*}

\begin{table*}
	\center
	\caption{Accuracy (\%) on Office-Home (ResNet-50).}
	\label{tab:office_home}
	\resizebox{\textwidth}{!}{
\addtolength{\tabcolsep}{-4pt} 
\begin{tabular}[htb]{cccccccccccccc}
\hline
Method & ${ar}\hspace{-1mm}\xrightarrow{}\hspace{-1mm}{cl}$ & ${ar}\hspace{-1mm}\xrightarrow{}\hspace{-1mm}{pr}$ & ${ar}\hspace{-1mm}\xrightarrow{}\hspace{-1mm}{rw}$ & ${cl}\hspace{-1mm}\xrightarrow{}\hspace{-1mm}{ar}$ &  ${cl}\hspace{-1mm}\xrightarrow{}\hspace{-1mm}{pr}$ & ${cl}\hspace{-1mm}\xrightarrow{}\hspace{-1mm}{rw}$ & ${pr}\hspace{-1mm}\xrightarrow{}\hspace{-1mm}{ar}$ & ${pr}\hspace{-1mm}\xrightarrow{}\hspace{-1mm}{cl}$ & ${pr}\hspace{-1mm}\xrightarrow{}\hspace{-1mm}{rw}$ & ${rw}\hspace{-1mm}\xrightarrow{}\hspace{-1mm}{ar}$ & ${rw}\hspace{-1mm}\xrightarrow{}\hspace{-1mm}{cl}$ & ${rw}\hspace{-1mm}\xrightarrow{}\hspace{-1mm}{pr}$ & Avg. \\ \hline
ResNet-50 &34.9 & 50.0 & 58.0 & 37.4 & 41.9 & 46.2 & 38.5 & 31.2 & 60.4 & 53.9 & 41.2 & 59.9 & 46.1 \\
DAN & 43.6 & 57.0 & 67.9 & 45.8 & 56.5 & 60.4 & 44.0 & 43.6 & 67.7 & 63.1 & 51.5 & 74.3 & 56.3 \\
DANN & 45.6 & 59.3 & 70.1 & 47.0 & 58.5 & 60.9 & 46.1 & 43.7 & 68.5 & 63.2 & 51.8 & 76.8 & 57.6 \\
JAN & 45.9 & 61.2 & 68.9 & 50.4 & 59.7 & 61.0 & 45.8 & 43.4 & 70.3 & 63.9 & 52.4 & 76.8 & 58.3 \\
CDAN & 49.0 & 69.3 & 74.5 & 54.4 & 66.0 & 68.4 & 55.6 & 48.3 & 75.9 & 68.4 & 55.4 & 80.5 & 63.8 \\
CDAN+E & 50.7 & \textbf{70.6} & \textbf{76.0} & 57.6 & 70.0 & 70.0 & 57.4 & 50.9 & 77.3 & 70.9 & 56.7 & 81.6 & 65.8 \\
Rot & 50.4 & 67.8 & 74.6 & 58.7 & 66.7 & 67.4 & 55.7 & 52.4 & 77.5 & 71.0 & 59.6 & 81.2 & 65.3 \\ \hline
Ours & \textbf{51.7} & 69.0 & 75.4 & \textbf{60.4} & \textbf{70.3} & \textbf{70.7} & \textbf{57.7} & \textbf{53.3} & \textbf{78.6} & \textbf{72.2} & \textbf{59.9} & \textbf{81.7} & \textbf{66.7}\\
\hline
\end{tabular}
}

\end{table*}

The results on PACS are shown in Table~\ref{tab:pacs}. \textbf{Dial} and \textbf{DDiscovery} are domain discovery based methods, 
\textbf{CDAN} and \textbf{CDAN+E} are adversarial training based methods, and \textbf{JiGen}, \textbf{Jigsaw} and \textbf{Rot} are self-supervised DA methods. Our method got constantly better accuracies than baseline Rot on four DA tasks and also outperforms sate-of-the-art methods \textbf{CDAN} and \textbf{CDAN+E}. 

The results on Office-31 dataset based on ResNet-50 are reported in Table~\ref{tab:office}. Our method got comparable accuracies to the top performing methods and again outperforms baseline \textbf{Rot} by around $2$ percentage points. The adversarial training base methods, {\eg} \textbf{CDAN} and \textbf{CDAN+E} achieved best accuracies on Office dataset. However, as we will show later, these methods converge much slower than our method as they need to optimize minimax objective function.

Table~\ref{tab:image_clef} shows the results on the ImageCLEF-DA dataset. The tree domains in ImageCLEF-DA dataset have equal size and balanced in each category, which makes it easier for domain adaptation than Office-31 dataset. Though there is little room to improve on this dataset, our method still achieves best accuracy comparing to other state-of-the-art methods. Our method outperforms the baseline \textbf{Rot} by more than $3$ percentage points.

The results on Office-Home are reported in Table~\ref{tab:office_home}. The office-home dataset has four domains and more categories, the images in each domain are visually more dissimilar, making it the most difficult one for domain adaptation. Our proposed method outperforms all baselines on this dataset as well.

\subsection{Analysis of experiments}

\subsubsection{Convergence analysis}

\begin{figure}
		\centering
		\begin{minipage}[!t]{0.48\textwidth}
			\includegraphics[width=\textwidth]{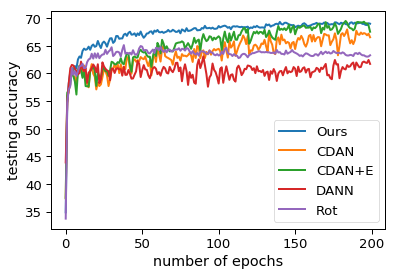}
			\caption{Convergence of different methods on task $D \rightarrow A$.}
			\label{fig:convergence}
		\end{minipage}
\end{figure}

We also empirically check the convergence of different DA methods, especially for comparison with adversarial based methods. \figurename~\ref{fig:convergence} shows the convergence of \textbf{DANN}, \textbf{CDAN}, \textbf{CDAN+E}, \textbf{Rot} and our proposed method on task $D \rightarrow A$. We run these methods with the same random seed and draw the testing accuracy on each epoch. \textbf{Rot} and our method show faster convergence and smoother curves than the other three adversarial based methods. The classical domain adversarial training method \textbf{DANN} has the lowest accuracy and it is also the most difficult one to converge. \textbf{CDAN} and \textbf{CDAN+E} improves the accuracy of \textbf{DANN}, but are still hard to converge due to the optimization of minimax objective function. From the curves, we can also see that when \textbf{Rot} gets plateau, the consistency training can still improve its accuracy.

\begin{figure}
	    \centering
		\begin{minipage}[!t]{0.48\textwidth}
			\includegraphics[width=\linewidth]{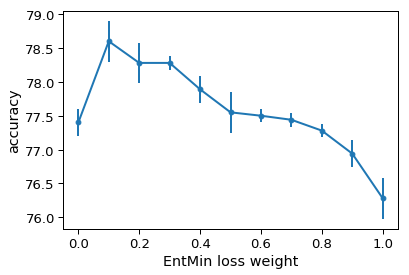}
			\caption{Ablation studies on ImageCLEF-DA dataset on task $I \rightarrow P$. The reported accuracy is the average over three repetitions. The error bar shows the value of standard deviation. We fix $\lambda_c = \lambda_p = 0.5$.} 
			\label{fig:ent_min}
		\end{minipage}
\end{figure}


\begin{table}
	\center
	\caption{Ablation studies on ImageCLEF-DA dataset on task $I \rightarrow P$.}
	\label{tab:ablation}
	 \begin{tabular}{ccccccc}
 \hline
 \multicolumn{7}{c}{$\lambda_p=0.6$, $\lambda_e=0.1$} \\
 $\lambda_c$ & 0.0   & 0.2  & 0.4  & 0.6  & 0.8  & 1.0 \\ \hline
 Avg.        & 77.9  & 78.6 & 78.1 & 77.9 & 77.7 & 78.0 \\
 std         & $\pm$0.8 & $\pm$0.4 & $\pm$0.4 & $\pm$0.4 & $\pm$0.5 & $\pm$0.6 \\ \hline \hline
 \multicolumn{7}{c}{$\lambda_c=0.2$, $\lambda_e=0.1$} \\
 $\lambda_p$ & 0.0   & 0.2  & 0.4  & 0.6  & 0.8  & 1.0 \\ \hline
 Avg.        & 78.3  & 78.3 & 78.5 & 78.6 & 78.5 & 78.3 \\
 std         & $\pm$0.1 & $\pm$0.4 & $\pm$0.4 & $\pm$0.4 & $\pm$0.4 & $\pm$0.4 \\
 \hline
 \end{tabular}
\end{table}

\begin{figure*}
	\centering
	\begin{minipage}[htb]{0.19\textwidth}
		\includegraphics[width=\textwidth]{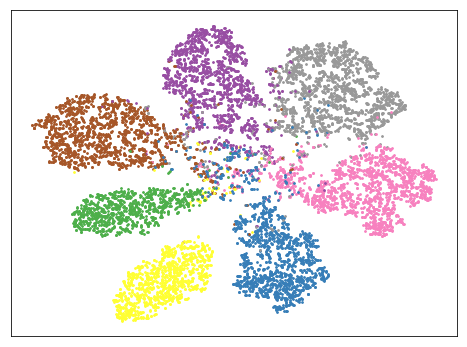}
		\centering \scriptsize
		(a)\ SRC
	\end{minipage}
	\begin{minipage}[htb]{0.19\textwidth}
		\includegraphics[width=\textwidth]{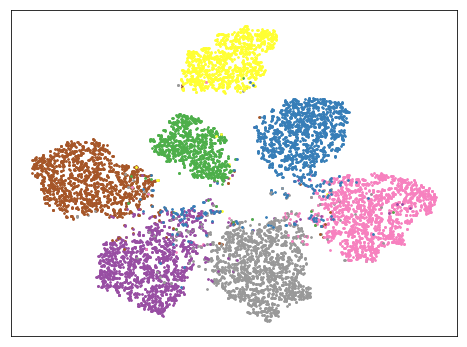}
		\centering \scriptsize
		(b)\ CDAN
	\end{minipage}
	\begin{minipage}[htb]{0.19\textwidth}
		\includegraphics[width=\textwidth]{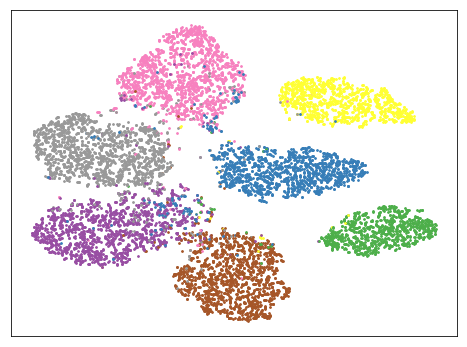}
		\centering \scriptsize
		(c)\ CDAN+E
	\end{minipage}
	\begin{minipage}[htb]{0.19\textwidth}
		\includegraphics[width=\textwidth]{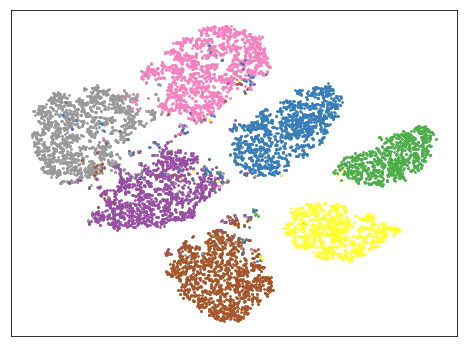}
		\centering \scriptsize
		(d)\ Rot
	\end{minipage}
	\begin{minipage}[htb]{0.19\textwidth}
		\includegraphics[width=\textwidth]{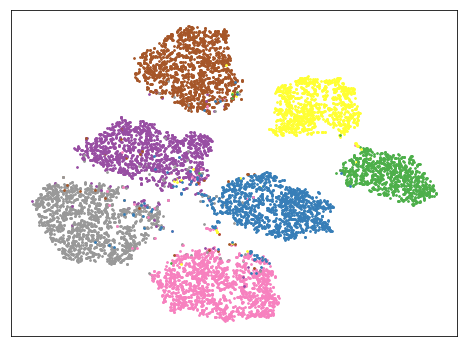}
		\centering \scriptsize
		(e)\ Ours
	\end{minipage}
	
	\begin{minipage}[htb]{0.19\textwidth}
		\includegraphics[width=\textwidth]{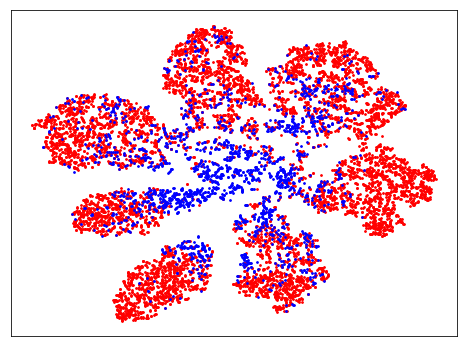}
		\centering \scriptsize
		(f)\ SRC
	\end{minipage}
	\begin{minipage}[htb]{0.19\textwidth}
		\includegraphics[width=\textwidth]{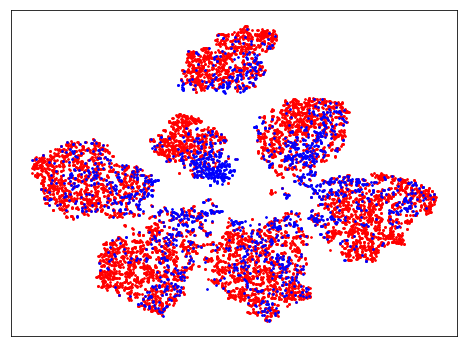}
		\centering \scriptsize
		(g)\ CDAN
	\end{minipage}
	\begin{minipage}[htb]{0.19\textwidth}
		\includegraphics[width=\textwidth]{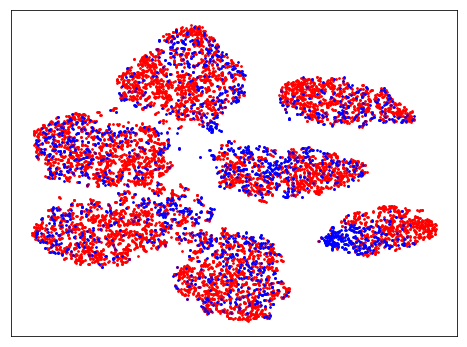}
		\centering \scriptsize
		(h)\ CDAN+E
	\end{minipage}
	\begin{minipage}[htb]{0.19\textwidth}
		\includegraphics[width=\textwidth]{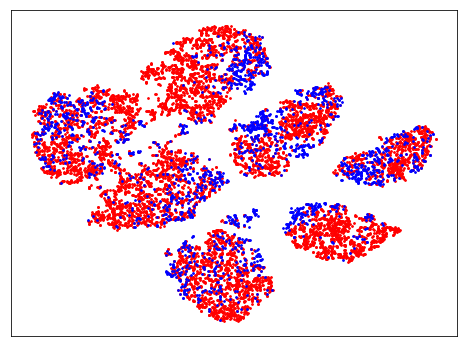}
		\centering \scriptsize
		(i)\ Rot
	\end{minipage}
	\begin{minipage}[htb]{0.19\textwidth}
		\includegraphics[width=\textwidth]{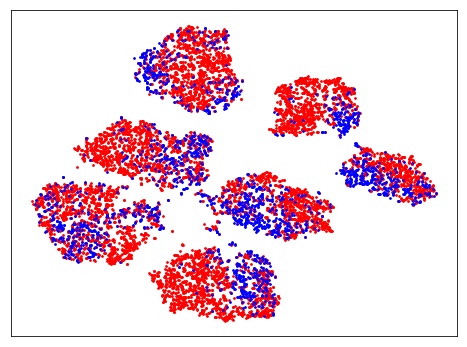}
		\centering \scriptsize
		(j)\ Ours
	\end{minipage}
	
	\caption{The t-SNE visualization of deep features in PACS DA task (art painting is used as target domain). (a)-(e) are feature distribution visualization with category colors. (f)-(j) are feature distribution visualization with domain colors. Red and blue points represent samples of source and target domains, respectively.}
	\label{fig:tsne-pacs}
\end{figure*}

\subsubsection{Ablation studies}

We perform ablation studies to the hyper-parameters on ImageCLEF-DA dataset. We first study the effect of conditional entropy minimization loss (EntMin) \Eq{eq:loss_ent_min}. For this experiment, we fix $\lambda_c = 0.5$ and $\lambda_p = 0.5$. \figurename~\ref{fig:ent_min} depicts the accuracies on task $I \rightarrow P$ with $\lambda_e$ ranging from $0.0$ to $1.0$.
When $\lambda_e = 0.0$, which means EntMin is excluded from the objective function, our method got accuracy around $77 .4\%$. With $\lambda_e = 0.1$, we obtained the best accuracy of $78.6\%$, which improves more than $1.0$ percentage point. When $\lambda_e$ is larger than $0.4$, the accuracy decreases. We keep $\lambda_e = 0.1$ for all our experiments as well as for baseline \textbf{Rot}.

Based on the results in \figurename~\ref{fig:ent_min}. Next, we fix $\lambda_e = 0.1$ and change the values of $\lambda_p$ and $\lambda_c$. Table~\ref{tab:ablation} shows the average accuracies and corresponding standard deviations with different hyper-parameters. In the first row, we fix $\lambda_e = 0.1$ and $\lambda_p = 0.6$, and change $\lambda_c$ from $0.0$ to $1.0$. It seems the DA accuracy is sensitive to $\lambda_c$ but vary in a short range. The best accuracy is at $\lambda_c = 0.2$. When fixing $\lambda_e = 0.1, \lambda_c = 0.2$ and changing $\lambda_p$, we obtained DA accuracies shown in the last two rows. It seems the proposed method is less sensitive to $\lambda_p$ than $\lambda_c$. We also found that the standard deviations are almost stable when changing $\lambda_c$ and $\lambda_p$, but sensitive to $\lambda_e$ as shown in \figurename~\ref{fig:ent_min}.

\subsubsection{Feature visualization}

Lastly, to better understand the learned cross-domain representations, we show t-SNE \cite{Maaten:2008} visualization of deep features in \figurename~\ref{fig:tsne-pacs}. Our method shows good discriminativity on categories and also very good domain alignment. The t-SNE visualization reveals that the proposed method learns domain invariant feature representations to achieve domain adaptation.

\section{Conclusion}

In this paper, we have shown that image rotation can be simultaneously used for both self-supervised learning and consistency training. By combining both, we have derived a principled way to handle the unlabeled data from target domain and thus have attained a new domain adaptation algorithm. 
The experimental results on multiple object recognition domain adaptation benchmarks have shown that consistency training constantly improves self-supervised domain adaptation. The evaluation on multiple domain adaptation datasets have shown that the proposed method achieves state-of-the-art performance. For future work, we would like to further explore the connections between self-supervised learning and consistency training and also consider to apply them for different tasks other than object recognition.






%

\bibliographystyle{IEEEtran}
\bibliography{main}

\begin{thebibliography}{10}
\providecommand{\url}[1]{#1}
\csname url@samestyle\endcsname
\providecommand{\newblock}{\relax}
\providecommand{\bibinfo}[2]{#2}
\providecommand{\BIBentrySTDinterwordspacing}{\spaceskip=0pt\relax}
\providecommand{\BIBentryALTinterwordstretchfactor}{4}
\providecommand{\BIBentryALTinterwordspacing}{\spaceskip=\fontdimen2\font plus
\BIBentryALTinterwordstretchfactor\fontdimen3\font minus
  \fontdimen4\font\relax}
\providecommand{\BIBforeignlanguage}[2]{{%
\expandafter\ifx\csname l@#1\endcsname\relax
\typeout{** WARNING: IEEEtran.bst: No hyphenation pattern has been}%
\typeout{** loaded for the language `#1'. Using the pattern for}%
\typeout{** the default language instead.}%
\else
\language=\csname l@#1\endcsname
\fi
#2}}
\providecommand{\BIBdecl}{\relax}
\BIBdecl

\bibitem{DAN:2015}
M.~Long, Y.~Cao, J.~Wang, and M.~I. Jordan, ``Learning transferable features
  with deep adaptation networks,'' in \emph{\ICML}, 2015.

\bibitem{rot:2019}
J.~{Xu}, L.~{Xiao}, and A.~M. {López}, ``Self-supervised domain adaptation for
  computer vision tasks,'' \emph{IEEE Access}, vol.~7, pp. 156\,694--156\,706,
  2019.

\bibitem{uda:2019}
Q.~Xie, Z.~Dai, E.~Hovy, M.~Luong, and Q.~V. Le, ``Unsupervised data
  augmentation for consistency training,'' \emph{arXiv}, 2019.

\bibitem{revisiting:2019}
A.~Kolesnikov, X.~Zhai, and L.~Beyer, ``Revisiting self-supervised visual
  representation learning,'' in \emph{\CVPR}, 2019.

\bibitem{doersch:2015}
C.~Doersch, A.~Gupta, and A.~A. Efros, ``Unsupervised visual representation
  learning by context prediction,'' in \emph{\ICCV}, 2015.

\bibitem{noroozi:2016}
M.~Noroozi and P.~Favaro, ``Unsupervised learning of visual representations by
  solving jigsaw puzzles,'' in \emph{\ECCV}, 2016.

\bibitem{gidaris:2018}
S.~Gidaris, P.~Singh, and N.~Komodakis, ``Unsupervised representation learning
  by predicting image rotations,'' in \emph{\ICLR}, 2018.

\bibitem{CSZ2006}
O.~Chapelle, B.~Schölkopf, and A.~Zien, Eds., \emph{Semi-Supervised
  Learning}.\hskip 1em plus 0.5em minus 0.4em\relax The MIT Press, 2006.

\bibitem{hu2017learning}
W.~Hu, T.~Miyato, S.~Tokui, E.~Matsumoto, and M.~Sugiyama, ``Learning discrete
  representations via information maximizing self-augmented training,'' in
  \emph{Proceedings of the 34th International Conference on Machine
  Learning-Volume 70}.\hskip 1em plus 0.5em minus 0.4em\relax JMLR. org, 2017,
  pp. 1558--1567.

\bibitem{CDAN:2018}
M.~Long, Z.~Cao, J.~Wang, and M.~I. Jordan, ``Conditional adversarial domain
  adaptation,'' in \emph{\NIPS}, 2018.

\bibitem{gan:2014}
I.~Goodfellow, J.~Pouget-Abadie, M.~Mirza, B.~Xu, D.~Warde-Farley, S.~Ozair,
  A.~Courville, and Y.~Bengio, ``Generative adversarial nets,'' in
  \emph{\NIPS}, 2014.

\bibitem{Gani:2015}
Y.~Gani, E.~Ustinova, H.~Ajakan, P.~Germain, H.~Larochelle, F.~Laviolette,
  M.~Marchand, and V.~Lempitsky, ``{Domain-Adversarial Training of Neural
  Networks},'' in \emph{\ICML}, 2015.

\bibitem{ADDA:2017}
E.~Tzeng, J.~Hoffman, K.~Saenko, and T.~Darrell, ``Adversarial discriminative
  domain adaptation,'' in \emph{\CVPR}, 2017.

\bibitem{JAN:2017}
M.~Long, H.~Zhu, J.~Wang, and M.~I. Jordan, ``Deep transfer learning with joint
  adaptation networks,'' in \emph{\ICML}, 2017.

\bibitem{carlucci:2019}
F.~M. Carlucci, A.~D'Innocente, S.~Bucci, B.~Caputo, and T.~Tommasi, ``Domain
  generalization by solving jigsaw puzzles,'' in \emph{\CVPR}, 2019.

\bibitem{Manasi06}
M.~Datar and X.~Qi, ``Automatic image orientation detection using the
  supervised self-organizing map,'' 2006, tech Report.

\bibitem{colorful:2016}
R.~Zhang, P.~Isola, and A.~A. Efros, ``Colorful image colorization,'' in
  \emph{\ECCV}, 2016.

\bibitem{pathak:2016}
D.~Pathak, P.~Krahenbuhl, J.~Donahue, T.~Darrell, and A.~A. Efros, ``Context
  encoders: Feature learning by inpainting,'' in \emph{\CVPR}, 2016.

\bibitem{jigsaw:2018}
D.~Kim, D.~Cho, D.~Yoo, and I.~S. Kweon, ``Learning image representations by
  completing damaged jigsaw puzzles,'' in \emph{WACV}, 2018.

\bibitem{Dosovitskiy:2014}
A.~Dosovitskiy, J.~T. Springenberg, M.~Riedmiller, and T.~Brox,
  ``Discriminative unsupervised feature learning with convolutional neural
  networks,'' in \emph{\NIPS}, vol.~1, 2014, pp. 766--774.

\bibitem{mixup:2018}
H.~Zhang, M.~Cisse, Y.~N. Dauphin, and D.~Lopez-Paz, ``mixup: Beyond empirical
  risk minimization,'' in \emph{\ICLR}, 2018.

\bibitem{vat:2018}
T.~Miyato, S.~ichi Maeda, S.~Ishii, and M.~Koyama, ``Virtual adversarial
  training: a regularization method for supervised and semi-supervised
  learning,'' \emph{\TPAMI}, 2018.

\bibitem{AutoAugment:2019}
E.~D. Cubuk, B.~Zoph, D.~Mane, V.~Vasudevan, and Q.~V. Le, ``Autoaugment:
  Learning augmentation strategies from data,'' in \emph{\CVPR}, June 2019.

\bibitem{pba:2019}
D.~Ho, E.~Liang, I.~Stoica, P.~Abbeel, and X.~Chen, ``Population based
  augmentation: Efficient learning of augmentation policy schedules,'' in
  \emph{\ICML}, 2019.

\bibitem{randaugment:2019}
E.~D. Cubuk, B.~Zoph, J.~Shlens, and Q.~V. Le, ``Randaugment: Practical data
  augmentation with no separate search,'' \emph{arXiv preprint
  arXiv:1909.13719}, 2019.

\bibitem{French:2018}
G.~French, M.~Mackiewicz, and M.~Fisher, ``Self-ensembling for visual domain
  adaptation,'' in \emph{\ICLR}, 2018, pp. 1--15.

\bibitem{EntMin:2005}
Y.~Grandvalet and Y.~Bengio, ``Semi-supervised learning by entropy
  minimization,'' in \emph{\NIPS}, 2005.

\bibitem{Deng:2009}
J.~Deng, W.~Dong, R.~Socher, L.-J. Li, K.~Li, and L.~Fei-Fei, ``Imagenet: A
  large-scale hierarchical image database,'' in \emph{\CVPR}, Miami, Florida,
  US, 2009.

\bibitem{pacs:2017}
D.~Li, Y.~Yang, Y.-Z. Song, and T.~M. Hospedales, ``Deeper, broader and artier
  domain generalization,'' in \emph{\ICCV}, 2017.

\bibitem{Saenko:2010}
K.~Saenko, B.~Hulis, M.~Fritz, and T.~Darrel, ``Adapting visual category models
  to new domains,'' in \emph{\ECCV}, 2010.

\bibitem{deephash:2017}
H.~Venkateswara, J.~Eusebio, S.~Chakraborty, and S.~Panchanathan, ``Deep
  hashing network for unsupervised domain adaptation,'' in \emph{\CVPR}, 2017.

\bibitem{carlucci:2017}
F.~M. Carlucci, L.~Porzi, B.~Caputo, E.~Ricci, and S.~R. Bulo, ``Just dial:
  Domain alignment layers for unsupervised domain adaptation,'' in
  \emph{International Conference on Image Analysis and Processing}, 2017.

\bibitem{mancini:2018}
M.~Mancini, L.~Porzi, S.~RotaBulo, B.~Caputo, and E.~Ricci, ``Boosting domain
  adaptation by discovering latent domains,'' in \emph{\CVPR}, 2018.

\bibitem{RTN:2016}
M.~Long, H.~Zhu, J.~Wang, and M.~I. Jordan, ``Unsupervised domain adaptation
  with residual transfer networks,'' in \emph{\NIPS}, 2016.

\bibitem{Maaten:2008}
L.~van~der Maaten and G.~Hinton, ``Visualizing data using t-sne,''
  \emph{Journal of Machine Learning Research}, 2008.

\end{thebibliography}

%
%

\end{document}